\pdfoutput=1
\documentclass[graybox, envcountchap]{svmult}

\usepackage{mathptmx}        
\usepackage{helvet}          
\usepackage{courier}         

\usepackage{makeidx}         
\usepackage{graphicx}        
\usepackage{multicol}        
\usepackage[bottom]{footmisc}

\usepackage{bbm}
\usepackage{svg}
\usepackage{siunitx}
\usepackage{wrapfig}
\usepackage{caption}
\usepackage{color}
\usepackage{siunitx}
\usepackage{placeins}

\usepackage{amsmath}
\usepackage{amsfonts}
\usepackage{amssymb}

\usepackage[normalem]{ulem}

\newcommand{\etal}{\textit{et al}.}

\newcommand{\eg}{\textit{e}.\textit{g}. }

\newcommand{\snuggg}{\vspace{-3em}}
\newcommand{\snugg}{\vspace{-2em}}
\newcommand{\snug}{\vspace{-1em}}
\newcommand{\push}{\vspace{1em}}

\newcommand*\samethanks[1][\value{footnote}]{\footnotemark[#1]}

\makeindex             

\begin{document}

\title*{Scaling simulation-to-real transfer by learning composable robot skills}
\author{
  Ryan Julian\thanks{Equal contribution},
  Eric Heiden\samethanks,
  Zhanpeng He,
  Hejia Zhang,\\
  Stefan Schaal,
  Joseph J. Lim,
  Gaurav Sukhatme,
  and Karol Hausman
}

\authorrunning{
R. Julian, E. Heiden, Z. He, H. Zhang, \etal 
}
\institute{
R. Julian, E. Heiden, Z. He, H. Zhang, S. Schaal, J.J. Lim, and G. Sukhatme
\at University of Southern California, Los Angeles, CA,\\ \email{\{rjulian, heiden, zhanpenh, hejiazha, sschaal, limjj, gaurav\}@usc.edu} \\
K. Hausman \at Google Brain, Mountain View, CA, \email{karolhausman@google.com}
}
%
%

\maketitle
\vspace{-2.5cm}
\abstract{
We present a novel solution to the problem of simulation-to-real transfer, which builds on recent advances in robot skill decomposition. 
Rather than focusing on minimizing the simulation-reality gap, we learn a set of diverse policies that are parameterized in a way that makes them easily reusable. 
This diversity and parameterization of low-level skills allows us to find a transferable policy that is able to use combinations and variations of different skills to solve more complex, high-level tasks.
In particular, we first use simulation to jointly learn a policy for a set of low-level skills, and a ``skill embedding'' parameterization which can be used to compose them. Later, we learn high-level policies which actuate the low-level policies via this skill embedding parameterization. The high-level policies encode how and when to reuse the low-level skills together to achieve specific high-level tasks.
Importantly, our method learns to control a real robot in joint-space to achieve these high-level tasks with little or no on-robot time, despite the fact that the low-level policies may not be perfectly transferable from simulation to real, and that the low-level skills were not trained on any examples of high-level tasks.
We illustrate the principles of our method using informative simulation experiments. We then verify its usefulness for real robotics problems by learning, transferring, and composing free-space and contact motion skills on a Sawyer robot using only joint-space control. We experiment with several techniques for composing pre-learned skills, and find that our method allows us to use both learning-based approaches and efficient search-based planning to achieve high-level tasks using only pre-learned skills.
}

\snugg
\section{Introduction}
\snug
\subsubsection*{Motivation}
\label{sec:motivation}
\snug
The Constructivist hypothesis proposes that humans learn to perform new behaviors by using what they already know~\cite{cooper1993paradigm}. 
To learn new behaviors, it proposes that humans leverage their prior experiences across behaviors, and that they also generalize and compose previously-learned behaviors into new ones, rather than learning them from scratch \cite{drescher1991made}. Whether we can make robots learn so efficiently is an open question. Much recent work on robot learning has focused on ``deep'' reinforcement learning (RL), inspired by achievements of deep RL in continuous control~\cite{lillicrap2015ddpg} and game play domains~\cite{mnih2015dqn}. While recent attempts in deep RL for robotics are encouraging \cite{levine2016end, chebotar-hausman-zhang17icml, gu2017deep}, performance and generality on real robots remains challenging.

A major obstacle to widespread deployment of deep RL on real robots is data efficiency: most deep RL algorithms require millions of samples to converge~\cite{duan2016benchmark}. Learning from scratch using these algorithms on a real robot is therefore a resource-intensive endeavor, \eg by requiring multiple robots to learn in parallel for weeks~\cite{levine2018robotarmy}. One promising approach is to train deep RL algorithms entirely in faster-than-real-time simulation, and transfer the learned policies to a real robot.
\snugg
\subsubsection*{Problem Statement}
\label{sec:problem_statement}
\snug
Our contribution is a method for exploiting hierarchy, while retaining the flexibility and expressiveness of end-to-end RL approaches.

Consider the illustrative example of block stacking. One approach is to learn a single monolithic policy which, given any arrangement of blocks on a table, grasps, moves, and stacks each block to form a tower. This formulation is succinct, but requires learning a single sophisticated policy. We observe that block stacking--and many other practical robotics tasks--is easily decomposed into a few reusable primitive skills (e.g. locate and grasp a block, move a grasped block over the stack location, place a grasped block on top of a stack), and divide the problem into two parts: learning to perform and mix the skills in general, and learning to combine these skills into particular policies which achieve high-level tasks.
\snugg
\subsubsection*{Related Work}
\label{sec:related_work}
\snug
Our approach builds on the work of Hausman~\etal~\cite{hausman2018learning}, that learns a latent space which parameterizes a set of motion skills, and shows them to be temporally composable using interpolation between coordinates in the latent space. In addition to learning reusable skills, we present a method which learns to compose them to achieve high-level tasks, and an approach for transferring compositions of those skills from simulation to real robots. Similar latent-space methods have been recently used for better exploration~\cite{abhishek-meta, diversity-is-all-you-need} and hierarchical RL~\cite{heess2016modulate, tuomas18latent-hrl, coreyes2018self}.

Our work is related to parameter-space meta-learning methods, which seek to learn a single shared policy which is easily generalized to all skills in a set, but do not address skill sequencing specifically. Similarly, unlike recurrent meta-learning methods, which implicitly address sequencing of a family of sub-skills to achieve goals, our method addresses generalization of single skills while providing an explicit representation of the relationship between skills. We show that explicit representation allows us to combine our method with many algorithms for robot autonomy, such as optimal control, search-based planning, and manual programming, in addition to learning-based methods. Furthermore, our method can be used to augment most existing reinforcement learning algorithms, rather than requiring the formulation of an entirely new family of algorithms to achieve its goals.

Previous works proposed frameworks such as Associative Skill Memories~\cite{pastor2012asm} and probabilistic movement primitives~\cite{rueckert2015movprim} to acquire a set of reusable skills.
Other approaches introduce particular model architectures for multitask learning, such as Progressive Neural Networks~\cite{rusu2016progressive} or Attention Networks~\cite{rajendran2017adaapt}.

Common approaches to simulation-to-real transfer learning include randomizing the dynamic parameters of the simulation~\cite{peng2017simreal}, and varying the visual appearance of the environment~\cite{sadeghi2017cadrl}. Another approach is explicit alignment: given a mapping of common features between the source and target domains, domain-invariant state representations~\cite{tzeng2015adaption}, or priors on the relevance of input features~\cite{kroemer2016mlp}, can further improve transferability. 

Our method can leverage these techniques to improve the stability of the transfer learning process in two ways: (1) by training transferable skills which generalize to nearby skills from the start and (2) by intentionally learning {\it composable} parameterizations of those skills, to allow them to be easily combined before or after transfer.

\snug
\section{Technical Approach}
\label{sec:approach}
\snugg
\begin{figure}
    \snuggg
    \sidecaption[b]
    \includegraphics[height=4cm]{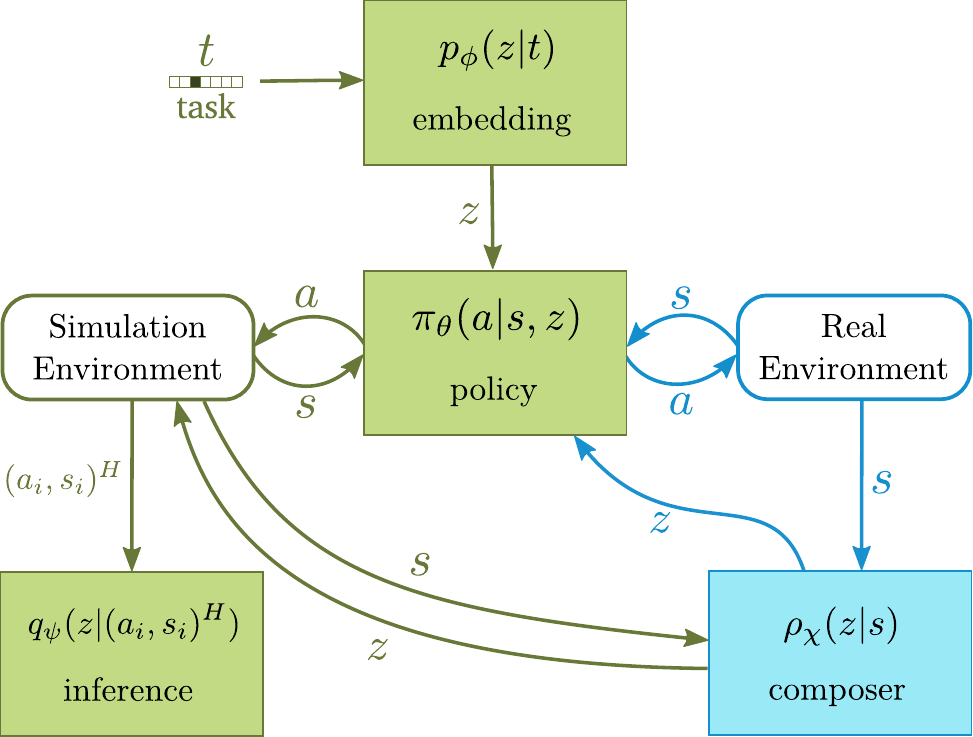}
    \caption{Block diagram of proposed architecture for transfer learning. Shared low-level skill components are shown in green. The high level task-specific component is shown in blue.}
    \label{fig:policy_architecture}
    \snug
\end{figure}

Our work synthesizes two recent methods in deep RL--pre-training in simulation and learning composable motion policies--to make deep reinforcement learning more practical for real robots. Our strategy is to split the learning process into a two-level hierarchy (Fig.~\ref{fig:policy_architecture}), with low-level skill policies learned in simulation, and high-level task policies learned or planned either in simulation or on the real robot, using the imperfectly-transferred low-level skills.

\snugg
\subsubsection*{Skill Embedding Learning Algorithm}
\label{sec:embedding_algo}
\snug

In our multi-task RL setting, we pre-define a set of low-level skills with IDs $T = \{ 1, \dots, N \}$, and accompanying, per-skill reward functions $r_{t \in T}(s, a)$.

In parallel with learning the joint low-level skill policy $\pi_\theta$ as in conventional RL, we learn an embedding function $p_\phi$ which parameterizes the low-level skill library using a latent variable $z$. Note that the true skill identity $t$ is hidden from the policy behind the embedding function $p_\phi$.
Rather than reveal the skill ID to the policy, once per rollout we feed the skill ID $t$, encoded as s one-hot vector, through the stochastic embedding function $p_\phi$ to produce a latent vector $z$. We feed this same value of $z$ to the policy for the entire rollout, so that all steps in a trajectory are correlated with the same value of $z$. 

\snugg
\begin{align}
    \label{eq:full_objective}
    \mathfrak{L}(\theta, \phi, \psi) &=
    \max_\pi \mathbb{E}_{\substack{\pi(a, z|s, t) \\ t\in T}}
        \left[
            \sum_{i=0}^\infty \gamma^i \hat{r}(s_i, a_i, z, t) \bigg| s_{i+1} \sim p(s_{i+1}|a_i, s_i)
        \right]
    \nonumber
    \\[-3.3em]
    &\intertext{where} \nonumber\\[-2.5em]
    \nonumber
    \hat{r}(s_i, a_i, z, t) &= \alpha_1 \mathbb{E}_{t\in T} \left[\mathbb{H}\{p_\phi(z|t)\}\right] + \alpha_2 \log q_\psi(z | s_i^H) + \alpha_3 \mathbb{H}\{\pi_\theta(a_i | s_i, z)\}
    \nonumber \\
    \nonumber
    &~~~ + r_t(s_i, a_i)  +  
    \nonumber
\end{align}
\snugg

To aid in learning the embedding function, we learn an inference function $q_\psi$ which, given a trajectory window $s_i^H$ of length $H$, predicts the latent vector $z$ which was fed to the low-level skill policy when it produced that trajectory. This allows us to define an augmented reward which encourages the policy to produce distinct trajectories for different latent vectors. We learn $q_\psi$ in parallel with the policy and embedding functions, as shown in Eq.~\ref{eq:full_objective}. 

We also add a policy entropy bonus $\mathbb{H}\{\pi_\theta(a_i | s_i, z)\}$, which ensures that the policy does not collapse to a single solution for each low-level skill, and instead encodes a variety of solutions.
All the above reward augmentations arise naturally from applying a variational lower bound to an entropy-regularized, multi-task RL formulation which uses latent variables as the task context input to the policy. For a detailed derivation, refer to~\cite{hausman2018learning}.

The full robot training and transfer method consists of three stages.

\snugg
\subsubsection*{Stage 1: Pre-Training in Simulation while Learning Skill Embeddings}
\snug
\label{sec:stage_1_pre_training}
We begin by training in simulation a multi-task policy $\pi_\theta$ for all low-level skills, and a composable parameterization of that library $p_\phi(z|t)$ (i.e. a skill embedding). This stage may be performed using any deep RL algorithm, along with the modified policy architecture and loss function described above. Our implementation uses Proximal Policy Optimization~\cite{schulman2017ppo} and the MuJoCo physics engine \cite{todorov2012mujoco}.

The intuition behind our pre-training process is as follows. The policy obtains an additional reward if the inference function is able to predict the latent vector which was sampled from the embedding function at the beginning of the rollout. This is only possible if, for every latent vector $z$, the policy produces a distinct trajectory of states $s_i^H$, so that the inference function can easily predict the source latent vector. Adding these criteria to the RL reward encourages the policy to explore and encode a set of diverse policies that can perform each low-level skill in various ways, parameterized by the latent vector.

\snugg
\subsubsection*{Stage 2: Learning Hierarchical Policies}
\label{sec:stage_2_learning}
\snug
In the second stage, we learn a high-level ``composer'' policy, represented in general by a probability distribution $\rho_\chi(z|s)$ over the latent vector $z$. The composer actuates the low-level policy $\pi_\theta(a | s, z)$ by choosing $z$ at each time step to compose the previously-learned skills. This hierarchical organization admits our novel approach to transfer learning: by freezing the low-level skill policy and embedding functions, and exploring only in the pre-learned latent space to acquire new tasks, we can transfer a multitude of high-level task policies derived from the low-level skills.

This stage can be performed directly on the the real robot or in simulation. As we show in Sec. \ref{sec:experiments}, composer policies may treat the latent space as either a discrete or continuous space, and may be found using learning, search-based planning, or even manual sequencing and interpolation. To succeed, the composer policy must explore the latent space of pre-learned skills, and learn to exploit the behaviors the low-level policy exhibits when stimulated with different latent vectors. We hypothesize that this is possible because of the richness and diversity of low-level skill variations learned in simulation, which the composer policy can exploit by actuating the skill embedding.

\snugg
\subsubsection*{Stage 3: Transfer and Execution}
\label{sec:stage_3_transfer}
\snug
Lastly, we transfer the low-level skill policy, embedding and high-level composer policies to a real robot and execute the entire system to perform high-level tasks.

\snugg
\section{Experiments}
\label{sec:experiments}

\snug
\subsubsection*{Point Environment}
\label{sec:point_environment}
\begin{figure}
    \snug
    \centering
    \includegraphics[height=3cm]{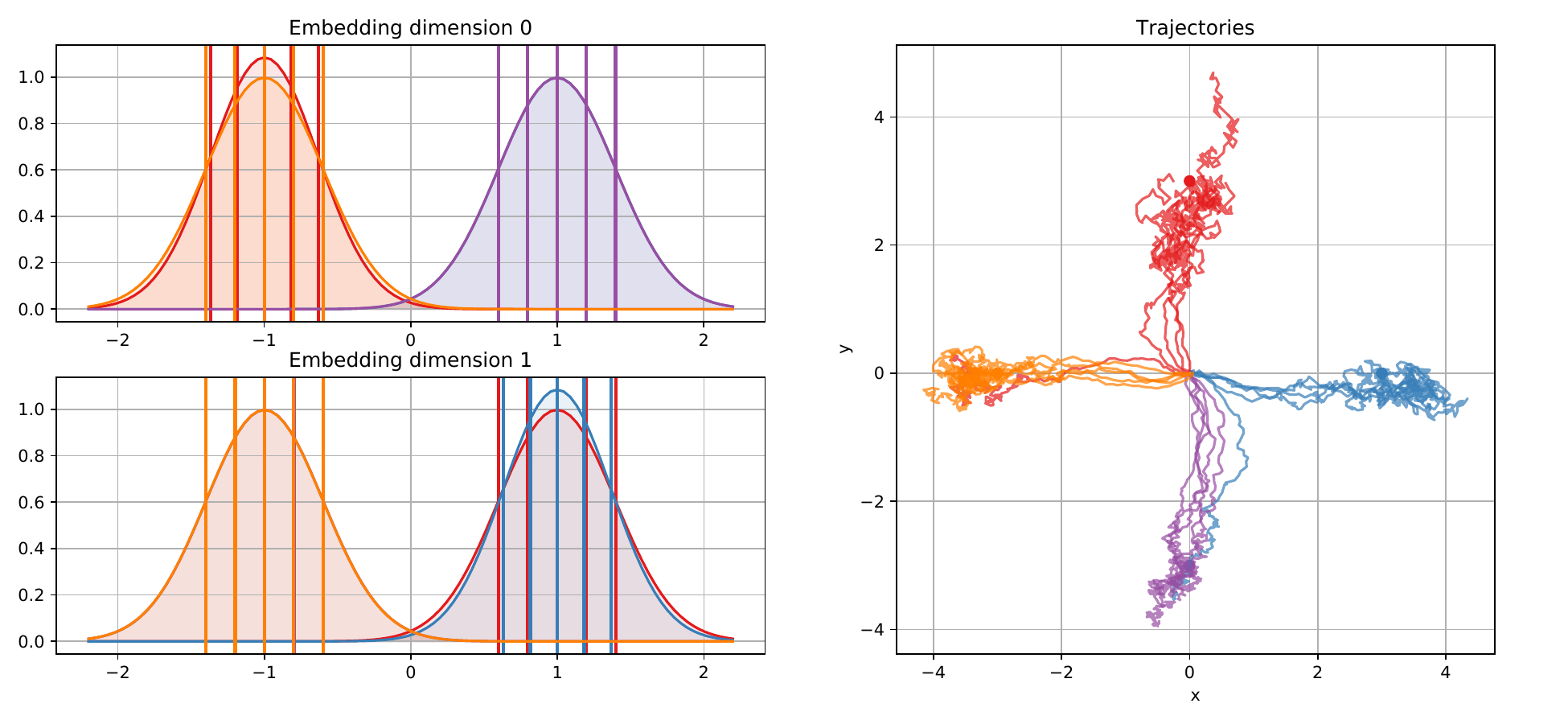}
    \captionof{figure}{Skill embedding distribution which successfully disentangles four different tasks using the embedding function.
    }
    \label{fig:point_embed}
    \snugg
\end{figure}

\snug
Before experimenting on complex robotics problems, we evaluate our approach in a point mass environment. Its low-dimensional state and action spaces, and high interpretability, make this environment our most basic test bed. We use it for verifying the principles of our method and tuning its hyperparameters before we deploy it to more complex experiments. Portrayed in Fig. \ref{fig:point_embed} is a multi-task instantiation of this environment with four goals (skills).

At each time step, the policy receives as state the point's position and chooses a two-dimensional velocity vector as its action. The policy receives a negative reward equal to the distance between the point and the goal position.

After 15,000 time steps, the embedding network learns a multimodal embedding distribution to represent the four tasks (Fig.~\ref{fig:point_embed}). Introducing entropy regularization~\cite{hausman2018learning} to the policy alters the trajectories significantly: instead of steering to the goal position in a straight line, the entropy-regularized policy encodes a distribution over possible solutions. Each latent vector produces a different solution. This illustrates that our approach is able to learn multiple distinct solutions for the same skill, and that those solutions are addressable using the latent vector input.

\snug
\subsubsection*{Sawyer Experiment: Reaching}
\label{sec:sawyer_experiment_reaching}
\begin{figure}
    \centering
    \includegraphics[height=3.5cm,trim=250pt 0 150pt 0,clip]{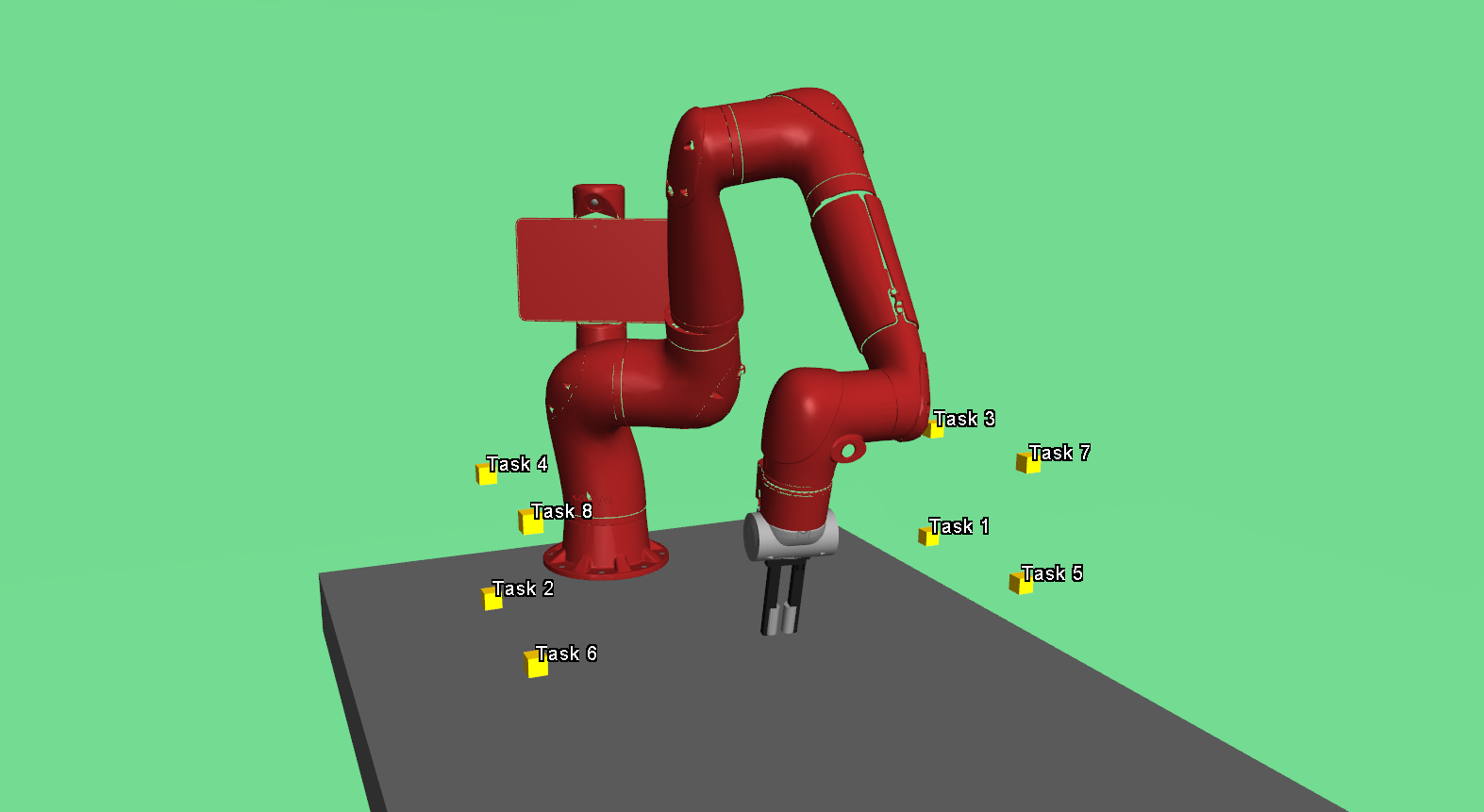}
    \includegraphics[height=3.5cm,trim=0 0 0 0,clip]{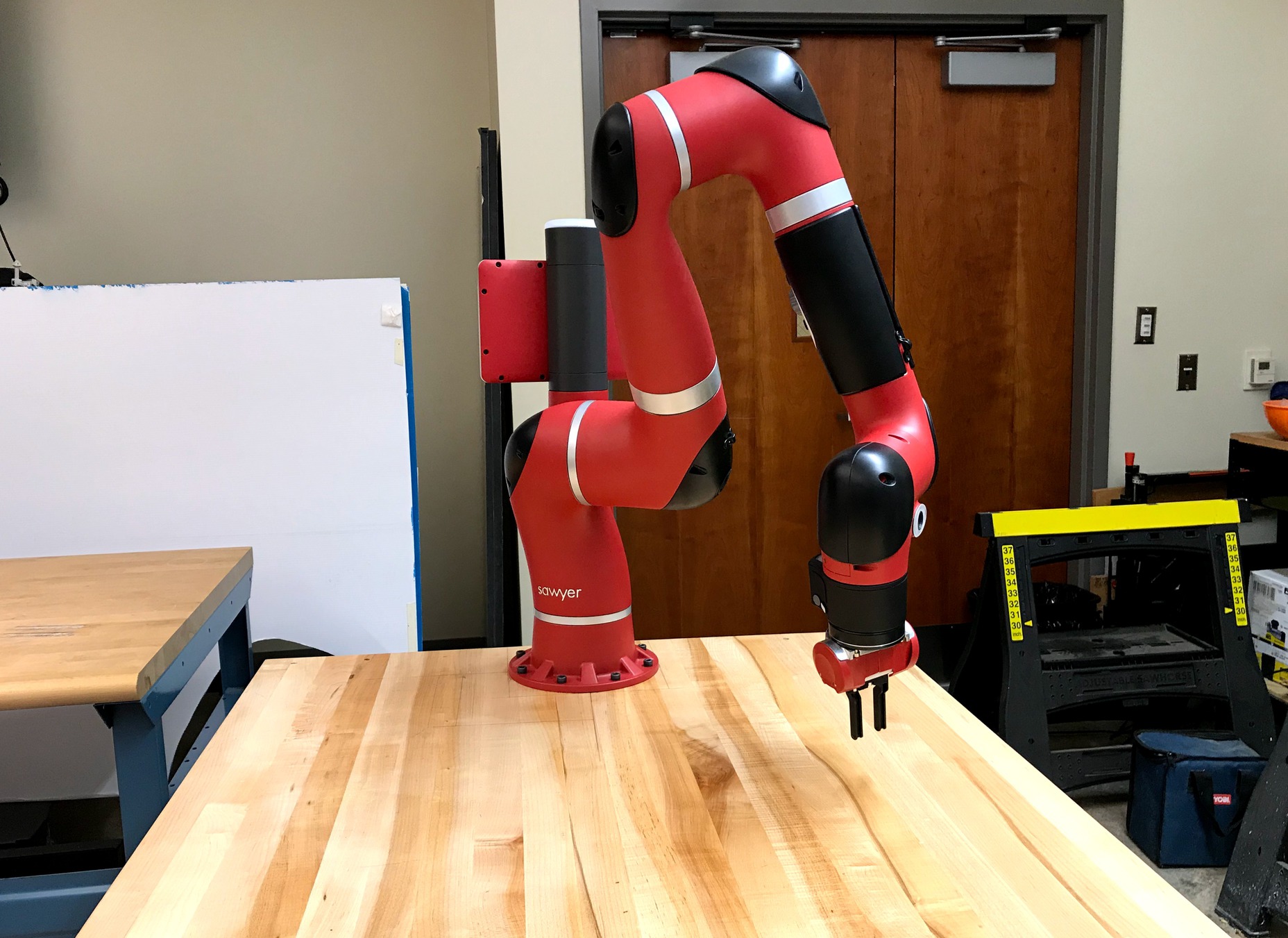}
    \caption{Multitask environment in simulation (left) and reality (right) for the reaching experiment. In pre-training, the robot learns a low-level skill for servoing its gripper to each of eight goal positions.}
    \label{fig:reach_envs}
    \snugg
\end{figure}

We ask the Sawyer robot to move its gripper to within \SI{5}{\cm} of a goal point in 3D space. The policy receives a state observation with the robot's seven joint angles, plus the cartesian position of the robot's gripper, and chooses incremental joint movements (up to \SI{0.04}{\radian}) as actions.

\begin{figure}
    \centering
      \includegraphics[height=3cm,trim=0 5pt 0 40pt,clip]{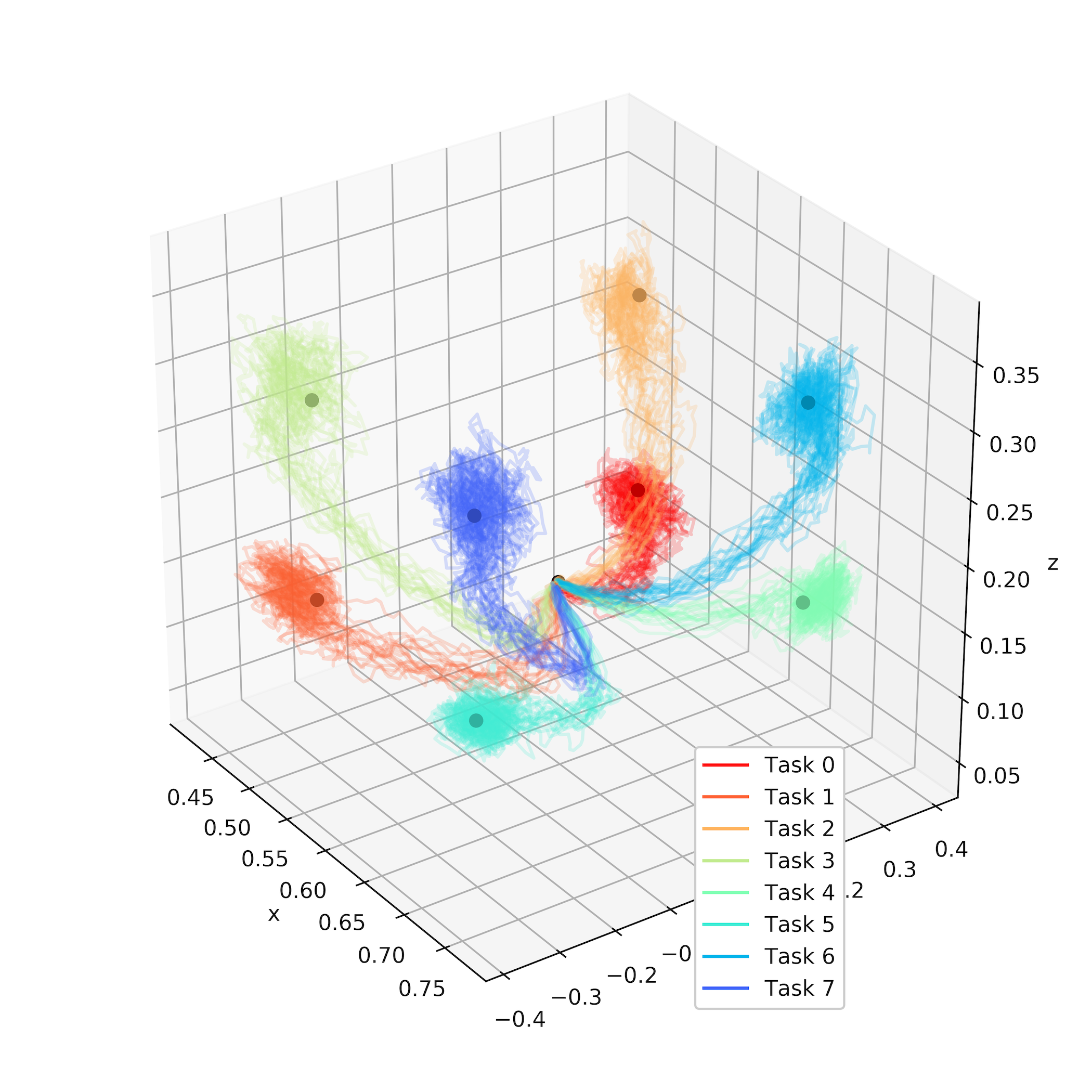}
    \includegraphics[height=3cm,trim=0 5pt 0 40pt,clip]{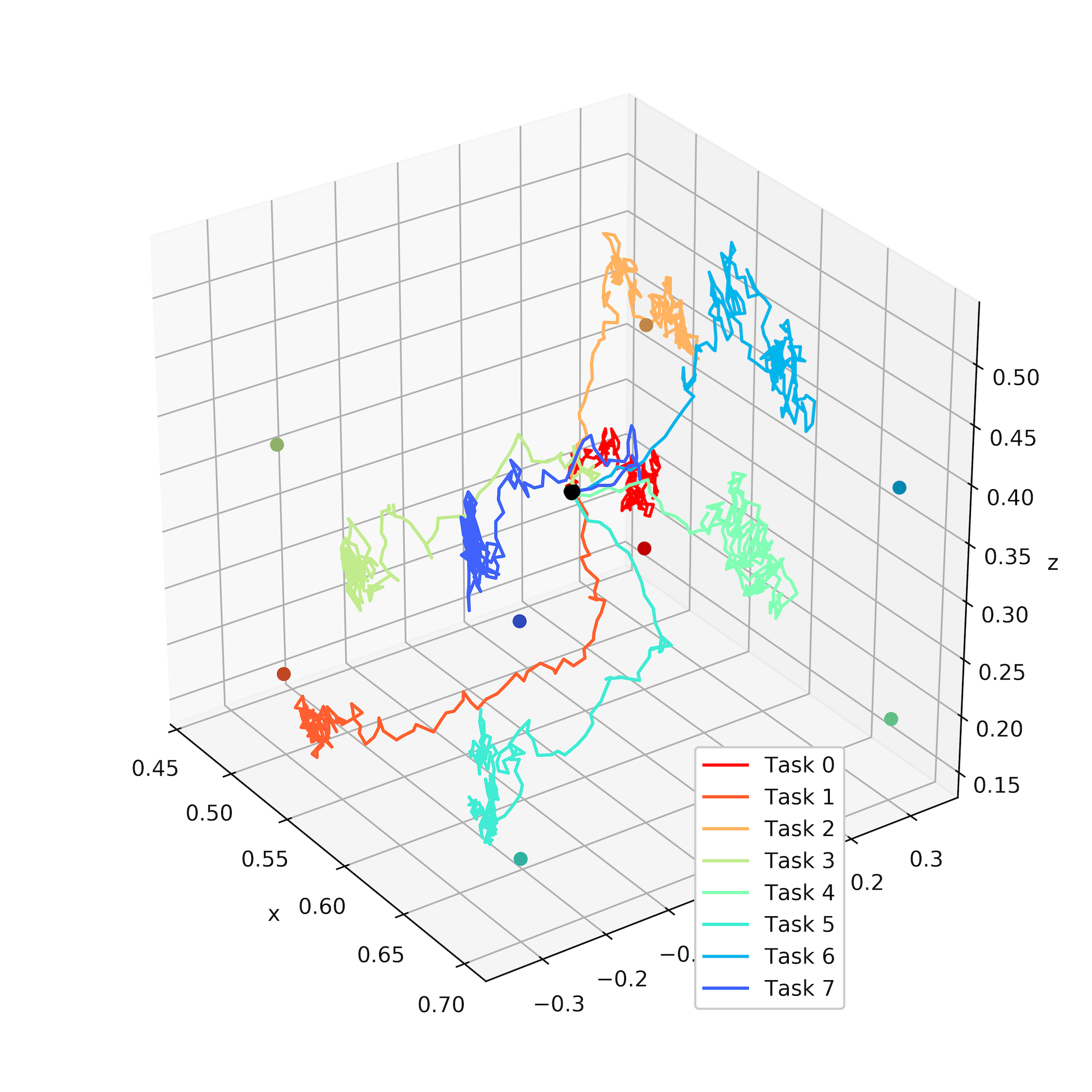}
    \caption{Gripper position trajectories for each skill from the low-level pre-trained reaching policy, after 100 training iterations. Left: simulation, right: real robot.}
    \label{fig:reacher_pos}
    \snug
\end{figure}

We trained the low-level policy on eight goal positions in simulation, forming a 3D cuboid enclosing a volume in front of the robot (Fig. \ref{fig:reacher_pos}). The composer policies feed latent vectors to the pre-trained low-level skill policy to achieve high-level tasks such as reaching previously-unvisited points (Fig. \ref{fig:reacher_compose}).

\snug

\snuggg
\subsubsection*{Composition Experiments}
\label{reacher_composition}
\snug

All Sawyer composition experiments use the same low-level skill policy, pre-trained in simulation. We experimented both with composition methods which directly transfer the low-level skills to the robot (direct), and with methods which use the low-level policy for a second stage of pre-training in simulation before transfer (sim2real).

\begin{figure}
    \centering
    \includegraphics[height=2.3cm, trim=0 0 0 0,clip]{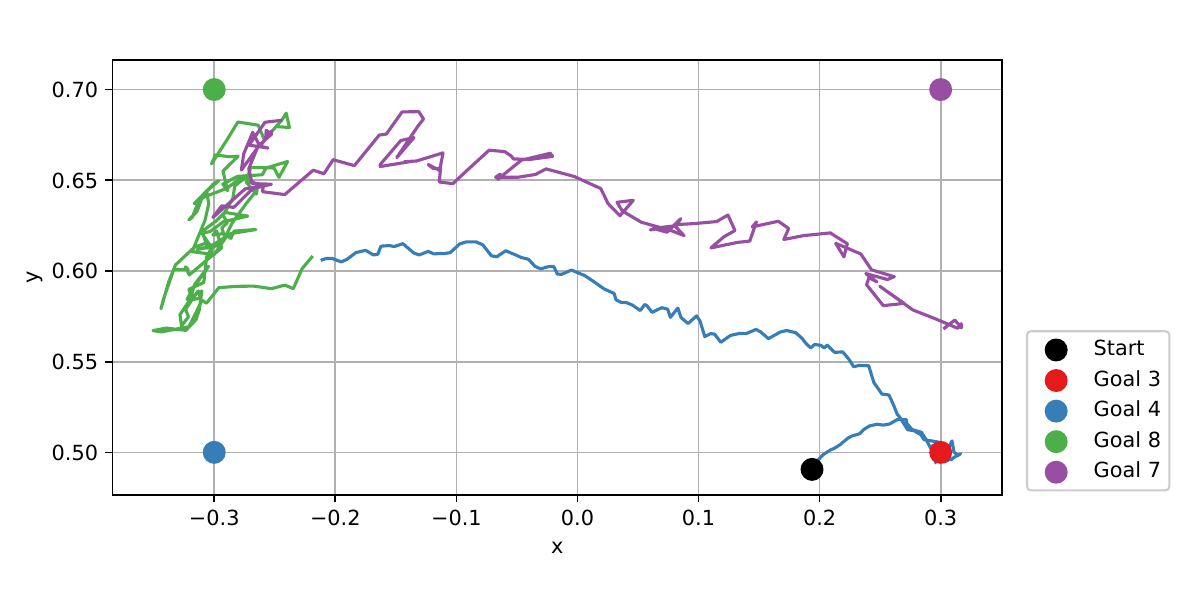}
    \includegraphics[height=2.8cm]{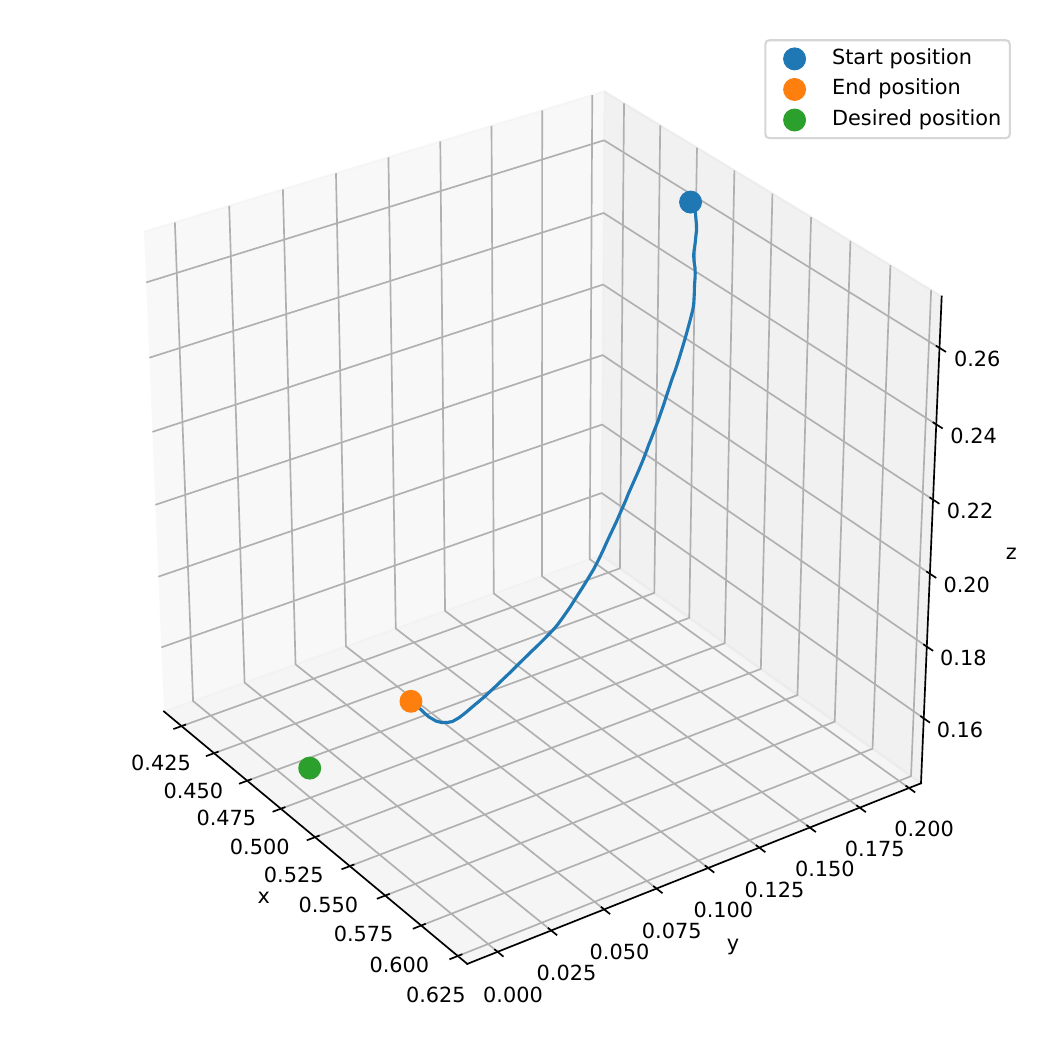}
    \includegraphics[height=2.8cm]{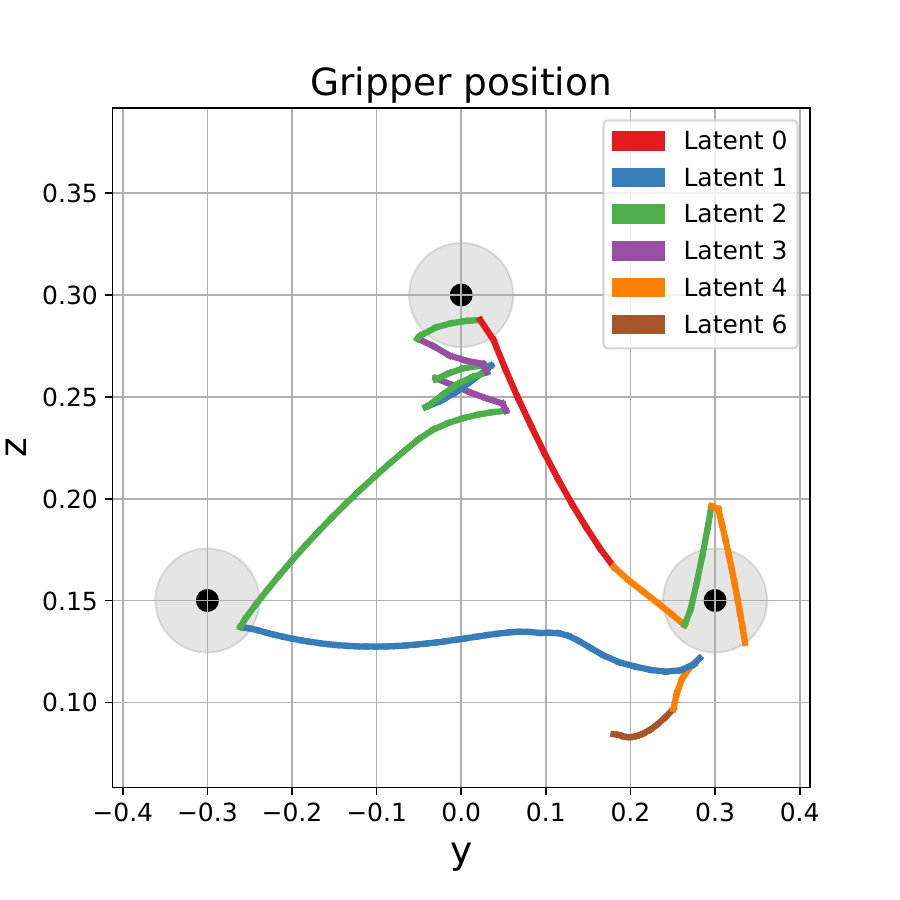}
    \caption{
        1. Gripper position trajectory while interpolating latents between task 3, 4, 8 and 7 for the embedded reaching skill policy on the real robot.
        2. Gripper position for composed reaching policy, trained with DDPG to reach an unseen point.
        3. Gripper position trajectory reaching points (black) in a triangle within a radius of \SI{5}{\cm} composed by search-based planning in the latent space.
    }
    \snugg
    \label{fig:reacher_compose}
\end{figure}

\noindent\textbf{Task interpolation in the latent space (direct)}

\noindent We evaluate the embedding function to obtain the mean latent vector for each of the 8 pre-training tasks, then feed linear interpolations of these means to the latent input of the low-level skill policy, transferred directly to the real robot. For a latent pair $(z_a, z_b)$, our experiment feeds $z_a$ for \SI{1}{\second}, then $z_i = \lambda_i z_a + (1 - \lambda_i)z_b \ \  \text{for} \ \  \lambda_i \in \left[0, 1\right]$ for \SI{1}{\second}, and finally $z_b$ for \SI{1}{\second}. We observe that the linear interpolation in latent space induces an implicit motion plan between the two points, despite the fact that pre-training never experienced this state trajectory. In one experiment, we used this method iteratively to control the Sawyer robot to draw a U-shaped path around the workspace (Fig.~\ref{fig:reacher_compose}.1).

\noindent\textbf{End-to-end learning in the latent space (sim2real)}

\noindent Using DDPG \cite{lillicrap2015ddpg}, an off-policy reinforcement learning algorithm, we trained a composer policy to modulate the latent vector to reach a previously-unseen point. We then transferred the composer and low-level policies to the real robot. The policy achieved a gripper distance error of \SI{5}{\cm}, the threshold for task completion as defined by our reward function (Fig.~\ref{fig:reacher_compose}.2).

\noindent\textbf{Search-based planning in the latent space (sim2real and direct)}

\noindent We used Uniform Cost Search in the latent space to find a motion plan (i.e. sequence of latent vectors) for moving the robot's gripper along a triangular trajectory. Our search space treats the latent vector corresponding to each skill ID as a discrete option. We execute a plan by feeding each latent in in the sequence to the low-level policy for $\sim$\SI{1}{\second}, during which the low-level policy executes in closed-loop.

In simulation, this strategy found a plan for tracing a triangle in less than \SI{1}{\minute}, and that plan successfully transferred to the real robot (Fig. \ref{fig:reacher_compose}.3). We replicated this experiment directly on the real robot, with no intermediate simulation stage. It took \SI{24}{\minute} of real robot execution time to find a motion plan for the triangle tracing task. 

\snug
\subsubsection*{Sawyer Experiment: Box Pushing}
\label{sec:sawyer_experiment_box_pushing}
\snug
\begin{figure}
    \snugg
    \centering
    \includegraphics[height=3.5cm,trim=100pt 0 150pt 0,clip]{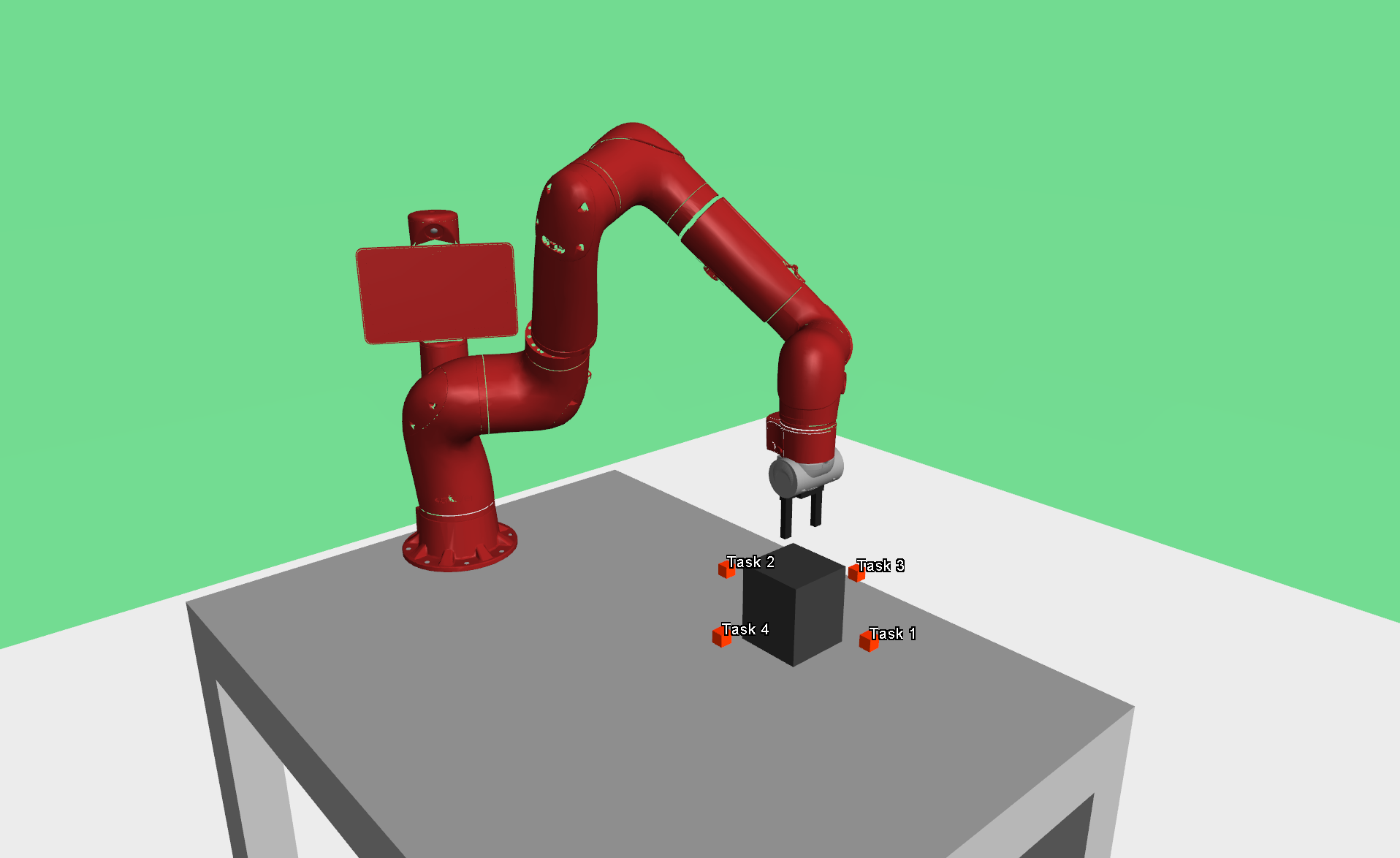}
    \includegraphics[height=3.5cm,trim=0 0 0 0, clip]{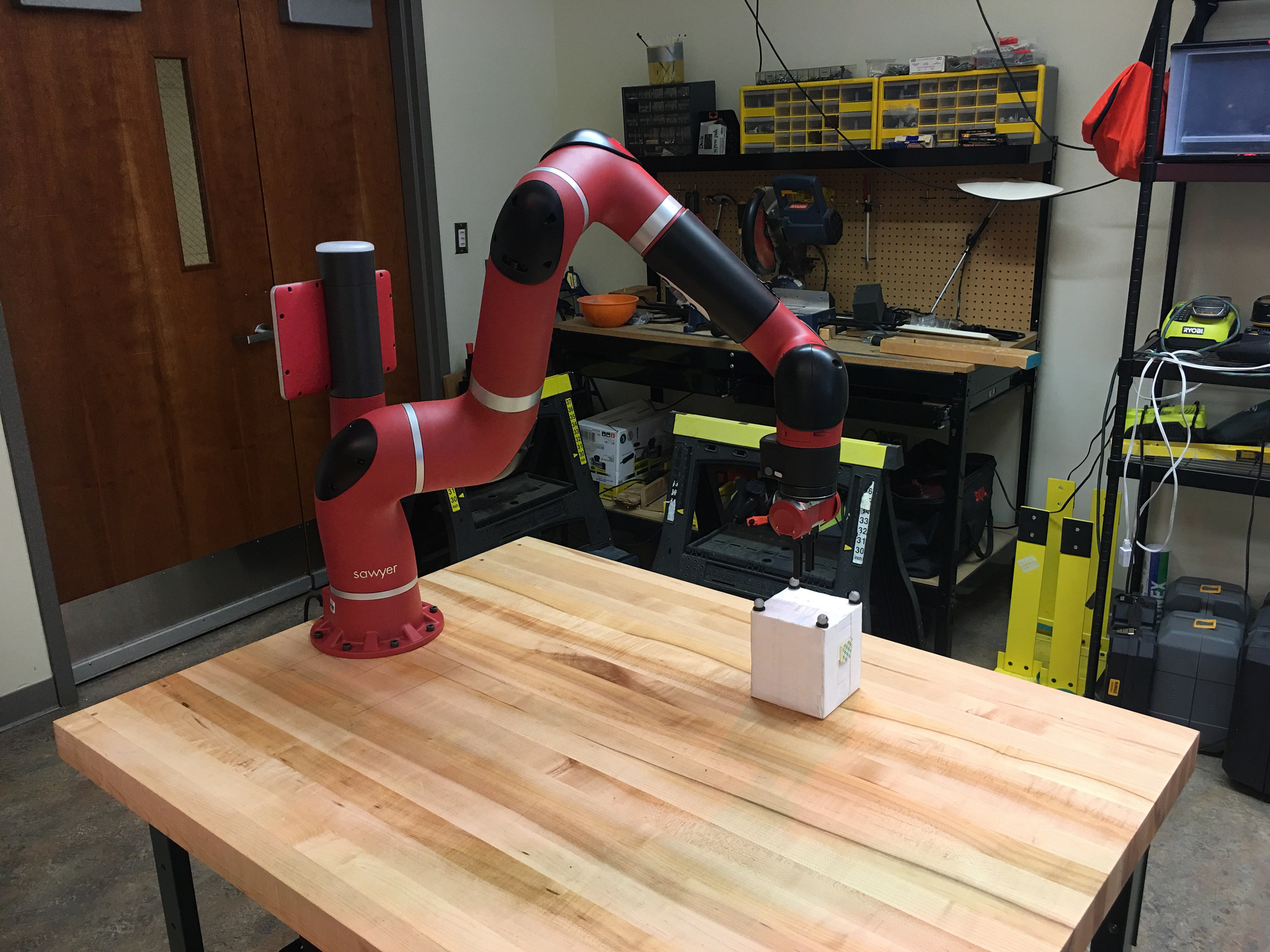}
    \caption{Multitask environment in simulation (left) and reality (right) for the box pushing experiment. Four goals are located on the table at a distance of \SI{15}{\cm} from the block's initial position.}
    \label{fig:push_multitask_env}
    \snugg
\end{figure}
We ask the Sawyer robot to push a box to a goal location relative to its starting position, as defined by a 2D displacement vector in the table plane. The policy receives a state observation with the robot's seven joint angles, plus a relative cartesian position vector between the robot's gripper and the box's centroid. The policy chooses incremental joint movements (up to $\pm$\SI{0.04}{\radian}) as actions. In the real experimental environment, we track the position of the box using motion capture and merge this observation with proprioceptive joint angles from the robot.

\begin{figure}
    \snug
    \centering
    \includegraphics[height=3cm]{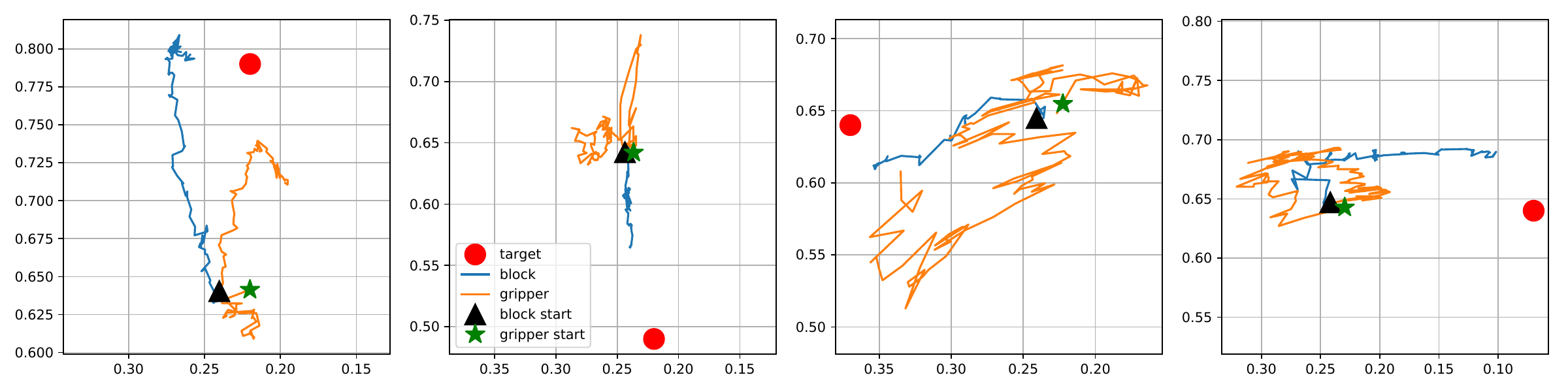}
    \caption{Trajectories of the block (blue) and gripper position (orange) created by execution of the embedded skill policy for each of the four pre-trained skills (red dots).}
    \label{fig:push_single_task}
    \snugg
\end{figure}

We trained the low-level pushing policy on four goal positions in simulation: up, down, left, and right of the box's starting point (Fig. \ref{fig:push_single_task}). The composer policies feed latent vectors to the pre-trained skill policy to push the box to positions which were never seen during training (Figs. \ref{fig:pusher_compose_means} and \ref{fig:pusher_compose_sequence}).

\snugg
\subsubsection*{Composition Experiments}
\snug
\noindent\textbf{Task interpolation in the latent space (direct)}
\begin{figure}
    \snug
    \centering
    \includegraphics[height=3cm]{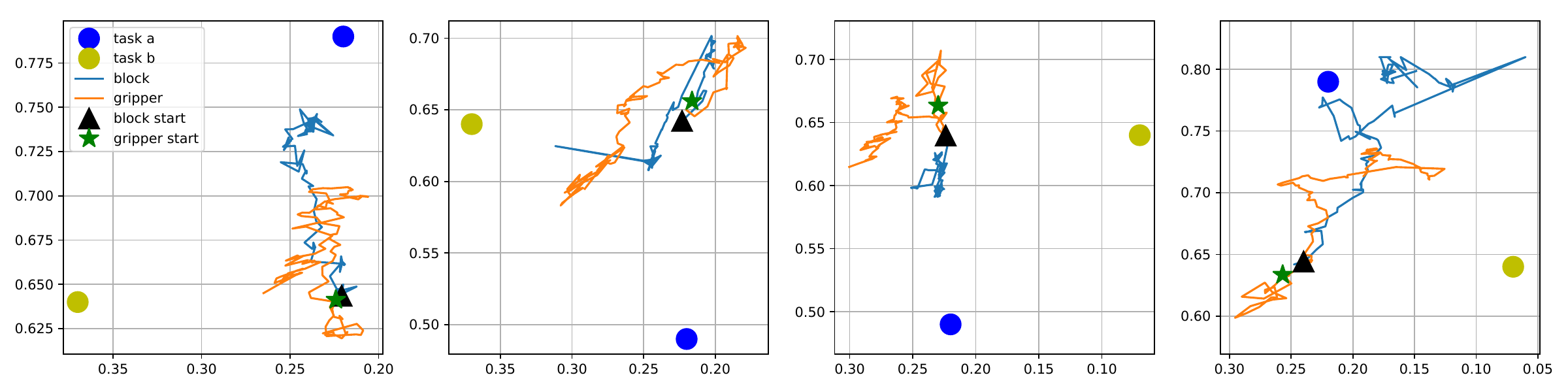}
    \caption{Trajectories of the block (blue) and gripper position (orange) created by executing the pre-trained embedded skill policy while feeding the mean latent vector of two neighboring skills. The robot pushes the block to positions between the goal locations of the two pre-trained skills.}
    \label{fig:pusher_compose_means}
    \snugg
\end{figure}

We evaluated the embedding function to obtain the mean latent vector for each of the four pre-trained pushing skills (i.e.\ up, down, left, and right of start position). We then fed the mean letant of adjacent skills (e.g.\ $z_\text{up-left} \sim mean\{z_\text{up}, z_\text{left}\}$) while executing the pre-trained policy directly on the robot (Fig. \ref{fig:pusher_compose_means}).

We find that in general this strategy induces the policy to move the block to a position between the two pre-trained skill goals. However, magnitude and direction of block movement was not easily predictable from the pre-trained goal locations, and this behavior was not reliable for half of the synthetic goals we tried.

\push
\noindent\textbf{Search-based planning in the latent space (sim2real)}
\begin{figure}
    \centering
    \snug
    \includegraphics[height=3cm]{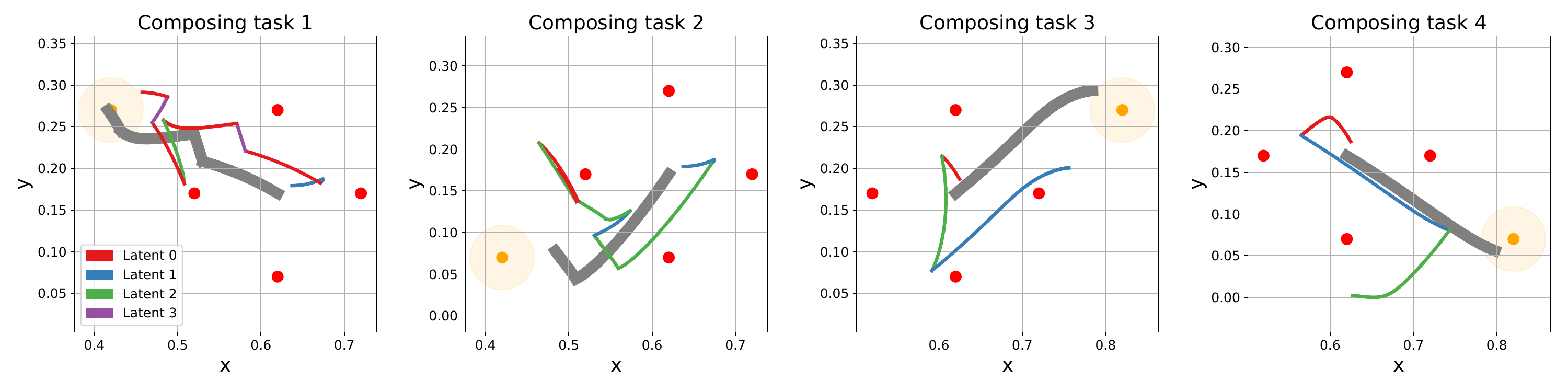}
    \caption{Search-based planning in the latent space achieves plans for pushing tasks where the goal position (orange dot) for the block is outside the region of tasks (red dots) on which the low-level skill was trained. Colored line segments trace the path of the robot gripper, and indicate which skill latent was used for that trajectory segment. The block's position is traced by the gray lines.}
    \label{fig:pusher_compose_sequence}
    \snug
\end{figure}

Similar to the search-based composer on the reaching experiment, we used Uniform Cost Search in the latent space to find a motion plan (sequence of latent vectors) for pushing the block to unseen goal locations (Fig.~\ref{fig:pusher_compose_sequence}). We found that search-based planning was able to find a latent-space plan to push the block to any location within the convex hull formed by the four pre-trained goal locations. Additionally, our planner was able to push blocks to some targets significantly outside this area (up to \SI{20}{\cm}). Unfortunately, we were not able to reliably transfer these composed policies to the robot.

We attribute these failures to transfer partially to the non-uniform geometry of the embedding space, and partially to the difficulty of transferring contact-based motion policies learned in simulation, and discuss these results further in Sec. \ref{sec:insights}.

\snug
\section{Main Experimental Insights}
\label{sec:insights}
The point environment experiments verify the principles of our method, and the single-skill Sawyer experiments demonstrate its applicability to real robotics tasks. Recall that all Sawyer skill policies used only joint space control to actuate the robot, meaning that the skill policies and composer needed to learn how using the robot to achieve task-space goals without colliding the robot with the world or itself.

The Sawyer composition experiments provide the most insight into the potential of latent skill decomposition methods for scaling simulation-to-real transfer in robotics. The method allows us to reduce a complex control problem--joint-space control to achieve task-space objectives--into a simpler one: control in latent skill-space to achieve task-space objectives.

We found that the method performs best on new skills which are interpolations of existing skills. We pre-trained on just eight reaching skills with full end-to-end learning in simulation, and all skills were always trained starting from the same initial position. Despite this narrow initialization, our method learned a latent representation which allowed later algorithms to quickly find policies which reach to virtually any goal inside the manifold of the pre-training goals. Composed policies were also able to  induce non-colliding joint-space motion plans between pre-trained goals (Fig. \ref{fig:reacher_compose}).

Secondly, a major strength of the method is its ability to combine with a variety of existing, well-characterized algorithms for robotic autonomy. In addition to model-free reinforcement learning, we successfully used manual programming (interpolation) and search-based planning on the latent space to quickly reach both goals and sequences of goals that were unseen during pre-training (Figs. \ref{fig:reacher_compose}, \ref{fig:pusher_compose_means}, and \ref{fig:pusher_compose_sequence}). Interestingly, we found that the latent space is useful for control not only in its continuous form, but also via a discrete approximation formed by the mean latent vectors of the pre-training skills. This opens the method to combination with large array of efficient discrete planning and optimization algorithms, for sequencing low-level skills to achieve long-horizon, high-level goals.

\begin{figure}
    \sidecaption[t]
    \includegraphics[height=3cm]{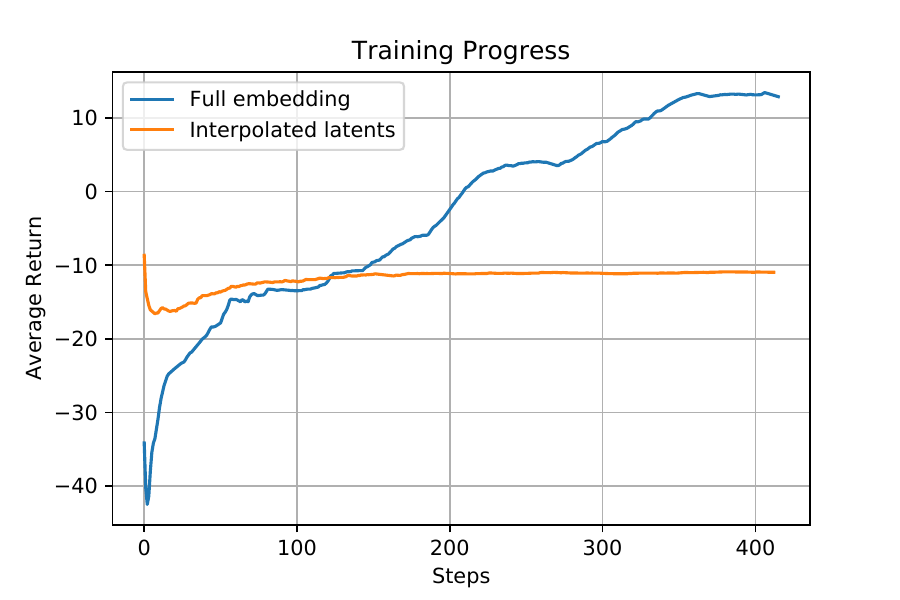}
    \caption{Training returns for a composing policy using an embedding trained on eight-goal reacher to reach a new point between the trained goals.}
    \label{fig:ddpg_compose_continuous_vs_discrete}
    \snugg
\end{figure}

Conversely, algorithms which operate on full continuous spaces can exploit the continuous latent space. We find that a DDPG-based composer with access only to a discrete latent space (formed from the latent means of eight pre-trained reaching skills and interpolations of those skills) is significantly outperformed by a DDPG composer that leverages the entire embedding space as its action space (Fig. \ref{fig:ddpg_compose_continuous_vs_discrete}). This implies that the embedding function contains information on how to achieve skills beyond the instantiations the skill policy was pre-trained on.

The method in its current form has two major challenges.

First is the difficulty of the simulation-to-real transfer problem even in the single-skill domain. We found in the Sawyer box-pushing experiment (Fig. \ref{fig:push_single_task}) that our ability to train transferable policies was limited by our simulation environment's ability to accurately model friction. This is a well-known weakness in physics simulators for robotics. A more subtle challenge is evident in Figure \ref{fig:reacher_pos}, which shows that our reaching policy did not transfer with complete accuracy to the real robot despite it being free-space motion task. We speculate that this is a consequence of the policy overfitting to the latent input during pre-training in simulation. If the skill latent vector provides all the information the policy needs to execute an open-loop trajectory to reach the goal, it is unlikely to learn closed-loop behavior.

The second major challenge is constraining the properties of the latent space, and reliably training good embedding functions, which we found somewhat unstable and hard to tune. The present algorithm formulation places few constraints on the algebraic and geometric relationships between different skill embeddings. This leads to counterintuitive results, such as the mean of two pushing policies pushing in the expected direction but with unpredictable magnitude (Fig. \ref{fig:pusher_compose_means}), or the latent vector which induces a reach directly between two goals (e.g A and B) actually residing much closer to the latent vector for goal A than for goal B (Fig. \ref{fig:reacher_compose}). This lack of constraints also makes it harder for composing algorithms to plan or learn in the latent space.

\snug
\section{Conclusion}
\snug
Our experiments illustrate the promise and challenges of applying of state-of-the-art deep reinforcement learning to real robotics problems. For instance, our policies were able to learn and generalize task-space control and even motion planning skills, starting from joint-space control, with no hand-engineering for those use cases. Simultaneously, the training and transfer process requires careful engineering and some hand-tuning. In the case of simulation-to-real techniques, our performance is also bounded by our ability to build and execute reasonable simulations of our robot and its environment, which is not a trivial task.

In future work, we plan to study how to learn skill embedding functions more reliably and with constraints which make them even more amenable to algorithmic composition, and further exploring how to learn and plan in latent space effectively. We also plan to combine our method with other transfer learning techniques, such as dynamics randomization~\cite{peng2017simreal}, to improve the transfer quality of embedded skill policies in future experiments. Our hope is to refine the method into a easily-applicable method for skill learning and reuse. We also look forward to further exploring the relationship between our method and meta-learning techniques, and the combination of our method with techniques for learning representations of the observation space.

\snugg
\section{Acknowledgements}
\label{sec:ack}
\snug
The authors would like to thank Angel Gonzalez Garcia, Jonathon Shen, and Chang Su for their work on the garage\footnote{\url{https://github.com/rlworkgroup/garage}} reinforcement learning for robotics framework, on which the software for this work was based.
This research was supported in part by National Science Foundation grants IIS-1205249, IIS-1017134, EECS-0926052, the Office of Naval Research, the Okawa Foundation, and the Max-Planck-Society. Any opinions, findings, and conclusions or recommendations expressed in this material are those of the author(s) and do not necessarily reflect the views of the funding organizations.

\bibliographystyle{unsrt}
\snugg
\bibliography{literature}

\end{document}